\pdfoutput=1
\documentclass[11pt]{article}

\usepackage[preprint]{acl}
\usepackage{times}
\usepackage{latexsym}
\usepackage{multirow}

\usepackage{tabularx}
\usepackage[T1]{fontenc}

\usepackage{devanagari}

\usepackage[utf8]{inputenc}

\usepackage{csquotes}
\usepackage{microtype}
\usepackage{graphicx}
\usepackage{subfig}

\usepackage{inconsolata}

\usepackage{enumitem}
\usepackage{graphicx}

\title{Domain-adaptative Continual Learning for Low-resource Tasks: Evaluation on Nepali}

\author{Sharad Duwal, Suraj Prasai, Suresh Manandhar \\
        Wiseyak\\\href{https://wiseyak.com/}{wiseyak.com}
        \\\\
 \small{
   \textbf{Corresponding Author:} \href{mailto:sharad.duwal@wiseyak.com}{sharad.duwal@wiseyak.com}
 }}

\begin{document}
\maketitle
\begin{abstract}
Continual learning has emerged as an important research direction due to the infeasibility of retraining large language models (LLMs) from scratch in the event of new data availability. Of great interest is the domain-adaptive pre-training (DAPT) paradigm, which focuses on continually training a pre-trained language model to adapt it to a domain it wasn’t originally trained on. In this work, we evaluate the feasibility of DAPT in a low-resource setting, namely the Nepali language. We use synthetic data to continue training Llama 3 8B to adapt it to the Nepali language in a 4-bit QLoRA setting. We evaluate the adapted model on its performance, forgetting, and knowledge acquisition. We compare the base model and the final model on their Nepali generation abilities, their performance on popular benchmarks, and run case-studies to probe their linguistic knowledge in Nepali. We see some unsurprising forgetting in the final model, but also surprisingly find that increasing the number of shots during evaluation yields better percent increases in the final model (as high as 19.29\% increase) compared to the base model (4.98\%), suggesting latent retention. We also explore layer--head self-attention heatmaps to establish dependency resolution abilities of the final model in Nepali. All code will be available at \href{https://github.com/sharad461/DAPT-Nepali}{github.com/sharad461/DAPT-Nepali}.
\end{abstract}

\section{Introduction}
Advancements in natural language processing (NLP) have enabled large language models (LLMs) to generate human-like text, follow instructions and perform well on a wide range of complex understanding tasks \cite{brown2020languagemodelsfewshotlearners, openai2024gpt4technicalreport, dubey2024llama3herdmodels}. 
A big driver behind the continued success of LLMs is the fact that scaling LLMs (increase in parameter count and dataset size) continues to provide decent returns on all performance benchmarks \cite{kaplan2020scalinglawsneurallanguage}. This scaling-up, however, affects the accessibility and availability of these models and comes with its myriad issues \cite{10.1145/3442188.3445922}. Very large language models require huge amounts of resources, have a large carbon footprint \cite{Strubell2019EnergyAP,Patterson2021CarbonEA}, and training them is feasible only for languages with large quantities of high-quality data and reasonable access to compute. It is costly also to perform inference on them.

Besides scaling, the other direction is generalizability of models with focus on optimal use of data. Given how human text data is projected to run out soon \cite{villalobos2024rundatalimitsllm}, methods like repeating data, using synthetic data, and using code data are being explored with good returns \cite{muennighoff2023scalingdataconstrainedlanguagemodels, shimabucoro2024llmseellmdo, aryabumi2024codecodeexploringimpact}. Many of these tools have been explored for research in low-resource languages.

Nepali is a low-resource language. \cite{arora-etal-2022-computational} classify Nepali among the "Scraping-By" languages in South Asia. While the frontier LLMs today can understand and generate Nepali \cite{openai2024gpt4technicalreport}, they do not officially support it. One major issue is tokenization: Nepali tokenization is costly in models like GPT4. While NLP research in South Asian languages has picked up recently, many languages are still behind and, as a result, low-resourced.

One possible way to ease the data-compute bind for these low-resource languages (Nepali included) is the use of continual learning (CL) for domain adaptation on high-resource LLMs \cite{sarvam, gururangan-etal-2020-dont}. The idea behind continual learning is to incrementally update an LLM with availability of new data so that the old knowledge isn’t forgotten and the new knowledge can be properly assimilated into the model. 

Domain adaptation with CL involves continued training of an LLM so that the knowledge of the base LLM can be repurposed to another domain. Since the \textit{knowledge} of the base model can be reused, we do not need large amounts of world knowledge data in the new domain (or language). Also, because adaptation requires training only a fraction of the total parameters in the original model, the compute requirements are significantly reduced.

In this work, we focus on domain adaptation of the Llama 3 8B \cite{llama3} model to the Nepali language using synthetically generated data. We continually train the Llama model, run experiments to determine performance, catastrophic forgetting, and linguistic knowledge acquisition of the model after the domain adaptation. We compare the adapted model against the original model on several benchmarks. Additionally, we analyze the attention heatmaps to gauge the knowledge of the adapted model. The emphasis of this work is on evaluating DAPT methods to adapt an LLM to a low-resource scenario with only synthetic data. 

The main contributions of this work are:
\begin{enumerate}[nosep]
    \item {We develop and test out methodologies to perform domain-adaptive continual pretraining on an open-weights model using only synthetically generated data.}
    \item {We evaluate and compare the performance of the adapted model against the base model.}
    \item {We interpret the linguistic knowledge of the final model on the new task.}
\end{enumerate}

\section{Related Work}
\textbf{Continual Learning.}
Continual learning is an important research direction because its goal is to make it possible to train large models on new data efficiently, often allowing lifelong learning LLMs. This could take place in the form of adding new information to it, teaching it a new subject, or adapting it to a different domain.

Domain-adaptive pretraining (DAPT) has been known to provide performance gains in low-resource settings \cite{gururangan-etal-2020-dont, yıldız2024investigatingcontinualpretraininglarge}. This has been extended to multilingual domain-adaptive pretraining where a single multilingual model is trained for a specific domain, which outperforms general models on said domain \cite{kaer-jorgensen-etal-2021-mdapt-multilingual}.

Synthetic data has also been applied for good performance gains in a continual pretraining domain-adaptation strategy \cite{zhang-etal-2020-multi-stage}.

However, a problem in continual learning is catastrophic forgetting, which happens during full finetuning probably due to retraining of weights or because a model has reached knowledge saturation and to learn any more information it forgets old information \cite{yıldız2024investigatingcontinualpretraininglarge}.

Continual learning has great potential in unlocking areas in low-resource language research.
\\\\
\textbf{Synthetic data.} Data augmentation using synthetic methods is central to research in low-resource languages. In NLP, some methods for synthetic data generation are backtranslation \cite{sennrich-etal-2016-improving}, paraphrasing, synonym replacement, sentence-level replacement, random insertion, etc. \cite{feng-etal-2021-survey} and \cite{chen-etal-2023-empirical} provide detailed studies on methods available for data augmentation for NLP tasks. 

Compared to real data, synthetic data has its own set of advantages and disadvantages. While synthetic data makes low-resource tasks accessible, scalable, and overall cost-effective, it might not always reflect realistic scenarios. There could often be challenges with validating synthetic data and it can magnify biases of the original model. 

Organic data available for training purposes is finite and \cite{villalobos2024rundatalimitsllm} predict we will run out of all publicly available text data as soon as 2026. Guided synthetic data generation, which will be an important part of future data acquisition technique, is a research direction where data is generated toward non-differentiable objectives \cite{shimabucoro2024llmseellmdo}.
\\\\
\textbf{Low-rank adaptation.} LoRA \cite{
hu2021loralowrankadaptationlarge} and QLoRA \cite{dettmers2023qloraefficientfinetuningquantized} are fine-tuning techniques that reduce the number of trainable parameters in a model, making training faster and memory-efficient. Instead of updating all weights in a model, these methods train low-rank matrices that capture task-specific information, freezing the model itself. In addition to the lower rank adaptation in LoRA, QLoRA quantizes the model so that it requires even lesser memory to train. The tradeoff in performance between full finetuning and low-rank techniques has been well-established \cite{biderman2024loralearnsforgets, xia2024chainloraefficientfinetuning}, and more work is being done in this space \cite{zhao2024galorememoryefficientllmtraining, lialin2023relorahighranktraininglowrank}, but in a resource-constrained scenario, QLoRA makes training large models feasible.
\\\\
\textbf{Knowledge in attention heads.} 
Many interpretative studies have been applied to the attention mechanism used in Transformer architectures. \cite{voita-etal-2019-analyzing} investigate the function of attention heads in the multi-head self-attention in encoders and try to interpret how they contribute to the performance of the entire network. They also prune attention heads in an ablation study. Similarly, to analyze how well the attention mechanism models a language and its syntax, \cite{vig-belinkov-2019-analyzing} evaluate attention heads to find that different layers in a model specialize in different parts-of-speech tags. They use BertViz \cite{vig-2019-multiscale} for their experiments. \cite{liu-etal-2019-linguistic} study the contextual representations generated by several popular models to understand why they are so effective in solving NLP tasks. They use seventeen probing tasks to establish the transferability of the representations and what linguistic knowledge is stored and in which part of the model.

\section{Method}
We use parallel data in Nepali--English (instead of Nepali-only text) to perform continual pretraining. Our aim here is to align the model and its knowledge to Nepali since it already has an understanding of English. We generate the parallel data using synthetic methods, perform pretraining on this data, and then finetune.

\subsection{Data Generation} \label{ssec:data}
We use Nepali text available online (news reports, essays, etc.) collected in datasets like OSCAR \cite{2022arXiv220106642A} and preprocess it for translation. For the translation system, there were a few alternatives to choose from: NLLB \cite{nllbteam2022languageleftbehindscaling}, IndicTrans2 \cite{gala2023indictrans2highqualityaccessiblemachine}, Google Cloud Translate. We use the Flores test-set for Nepali--English \cite{guzman-etal-2019-flores} to evaluate the open-source systems. We also compare scores across the different model sizes available and the various quantized versions of the models. We decided to go with 8-bit NLLB for the translation. IndicTrans2 performed marginally better in terms of BLEU scores, but NLLB had very little computational overhead and supported larger batches out-of-the-box.

Since we also plan to later finetune the model on Nepali instructions and since there aren’t instruction sets for Nepali, we also translate English instruction sets to Nepali. We use IndicTrans2 for this. For the instruction set, we translate Alpaca \cite{alpaca}, Databricks Dolly \cite{DatabricksBlog2023DollyV2} and WebGLM-QA \cite{liu2023webglm} to Nepali. To ensure the quality of the synthetic instruction sets, we backtranslate the instructions to English (again using IndicTrans2) and calculate the chrF++ score between the original and the backtranslated sets. We apply a chrF++ cut-off of $50$ and all samples with lower scores were discarded.

At the end of this step, we have 5M pairs of Nepali–English parallel paragraphs and 114K triplets of (input, instruction, output) instructions.

\subsection {Training}
We perform 4-bit QLoRA continual pretraining of a Llama 3 8B model on the synthetic parallel data we generated in \ref{ssec:data}. We use Unsloth \cite{unsloth}. We pretrain the model with the task to translate from English to Nepali. We do this because the English part of the parallel data is synthetic and the Nepali part is organic. 

We loosely follow the steps suggested by \cite{sarvam} and divide the pretraining process into two steps:

\subsubsection {Pretraining using translation} \label{sssec:pt}
The aim of this step is to familiarize the model with Nepali using the translation data and the model’s own knowledge in English. We train the model to translate from English to Nepali. We use this translation direction because for our parallel data, English is synthetic and Nepali is organic. By training the model to generate the (non-synthetic) Nepali given the (synthetic) English, we teach it to generate Nepali as originally written. The alternative would be to teach the model to generate system-generated English.

For this step, we set the rank to $128$, which selected ~335M parameters to train. We pretrain the model on 1.5M paragraph pairs for this first task.

\subsubsection {Bilingual next token prediction} \label{sssec:bnt}
Second, we train the model on a bilingual next token prediction task. This is the standard next token prediction task with sentences ordered in alternate language. We choose the next 1.5M paragraph pairs and consolidate each of the pairs such that every sample paragraph switches language every sentence. If the first sentence in a paragraph is Nepali, the second picks up in English, then back to Nepali. An example paragraph would be: 

\begin{displayquote}
Before the unification of Nepal, the Kathmandu Valley was known as Nepal. {\dn n\?pAl fNdko sVFk u(pE\381w aEnE\396wt C.}\ \ But it can be dated back to the fourth century AD.
\end{displayquote}

The training settings are much the same for this step as the first step. The presumption here is that instead of training with a Nepali next token prediction task, if we leverage the English knowledge already present into the model, the training should be more effective. \cite{sarvam} found that a model trained with this objective performed better than a model trained on the standard token prediction objective on 5X more data.

\subsection {Finetuning}
After these steps aligning the model (\ref{sssec:pt} and \ref{sssec:bnt}) to the Nepali language, we perform a supervised finetuning step. We perform a QLoRA finetuning lower-rank than both these pretraining steps. We set the rank to $16$. This updates around 41M parameters in the model. The instruction data we generated in \ref{ssec:data} is used to finetune the model here. We choose to perform finetuning on a mixed instruction set because we want the model to learn both Nepali and English instructions.

\section{Performance Study}
After the pretraining followed by finetuning, we perform experiments on both the base model (Llama 3 8B 4-bit) and the continual trained model with the view to answer the following research questions:

\begin{enumerate}[nosep]
    \item[Q1.]{Has the model learned Nepali?}
    \item[Q2.]{Has the model retained its knowledge of English? What does catastrophic forgetting look like?}
    \item[Q3.]{From a linguistic perspective, how well does the new model model the Nepali language?}
\end{enumerate}

\section{Experimental Setup}
\subsection{LM Evaluation Harness}
LM Evaluation Harness \cite{eval-harness} is a framework for evaluating language models. It supports generative LLMs trained on transformers, GPT-NeoX, and Megatron-DeepSpeed and as of writing it supports more than 60 academic benchmarks to run evaluations on. For our task, we focus on English benchmarks and evaluate first the base model, then the adapted model in order to quantify the change in model knowledge and performance.

\subsection{BertViz}
BertViz \cite{vig-2019-multiscale} is a tool designed to help visualize attention in language models. Originally designed to support only BERT-type models, decoder-only and encoder-decoder model support was added later. It provides a user-friendly interface to explore and interpret the attention patterns within the model, offering valuable insights into how LLMs process and relate different parts of the input with itself or with the output, facilitating in interpretative study of LLMs.

\begin{figure*}[!tb]
  \centering
  \subfloat[Scores for base model]{\includegraphics[width=0.5\textwidth]{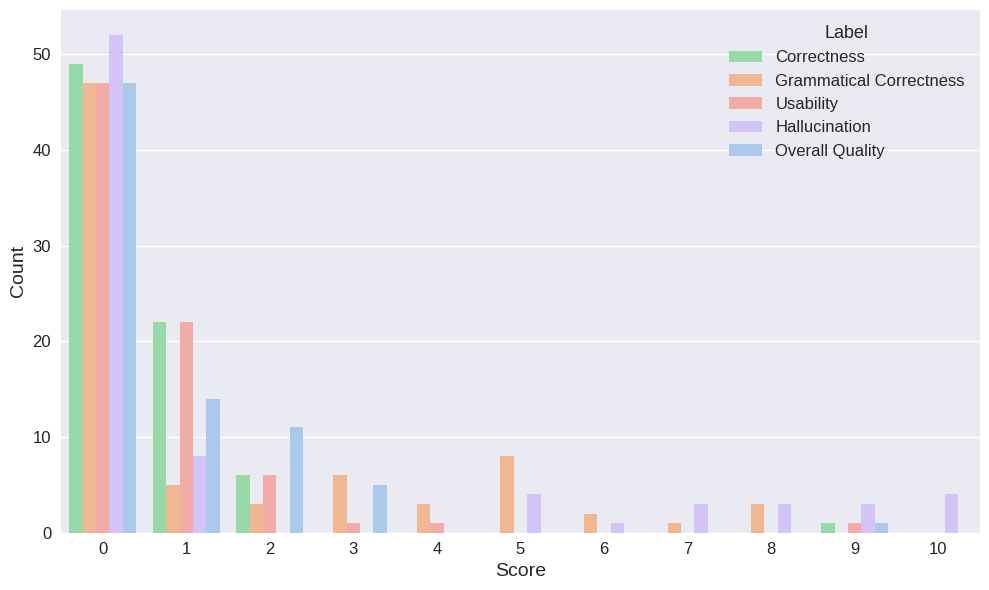}\label{fig:f1}}
    \hfill
  \subfloat[Scores for our model]{\includegraphics[width=0.5\textwidth]{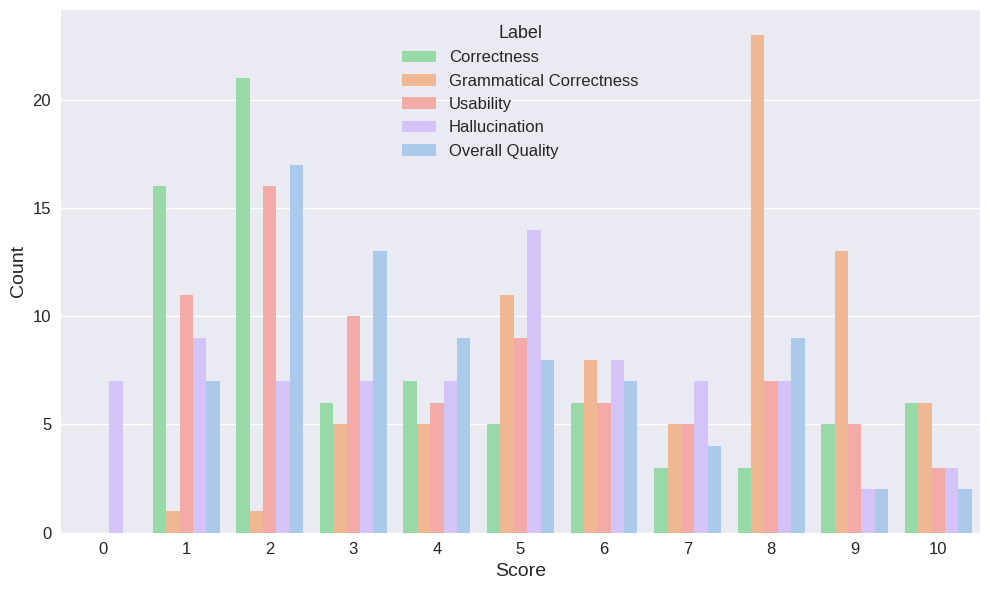}\label{fig:f2}}
    \vfill
  \subfloat[Base model]{\includegraphics[width=0.5\textwidth]{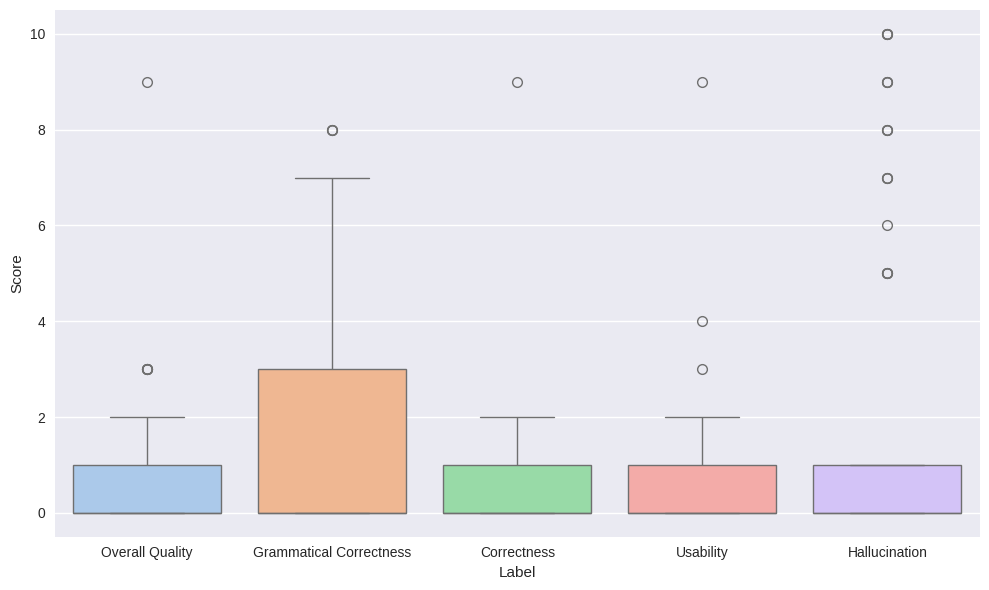}\label{fig:f3}}
    \hfill
  \subfloat[Our model]{\includegraphics[width=0.5\textwidth]{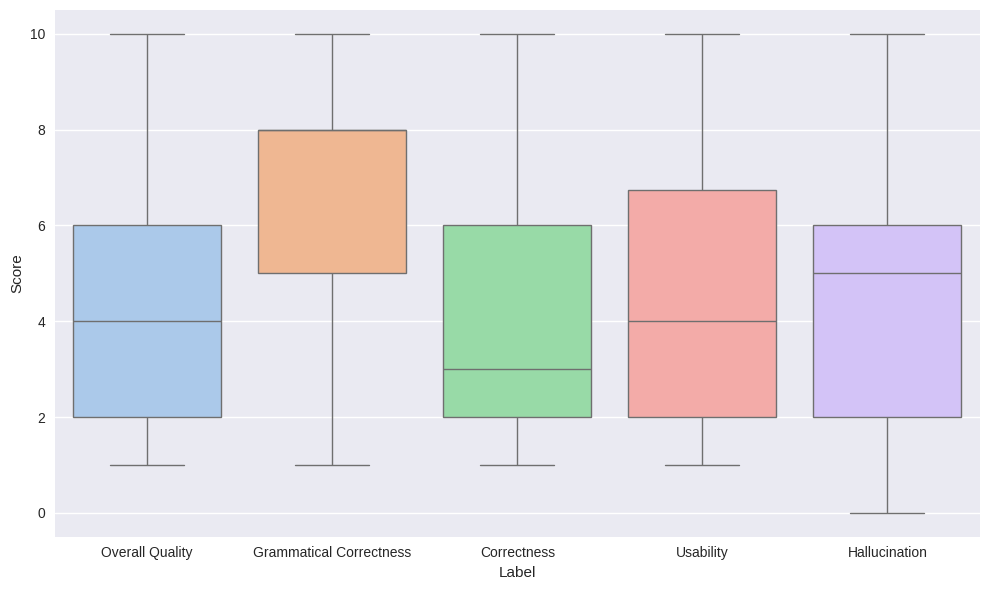}\label{fig:f4}}
  \caption{GPT4o scores for Nepali answers generated by the base model (Llama 3 8B 4-bit) and our model on five attributes: correctness, grammar, usability, hallucination and overall quality. Empty generations from the models are scored 0 on all attributes. c) and d) are the distribution of scores among the attributes with medians and outliers.}
\end{figure*}

\begin{table*}
  \centering
\begin{tabular}{ccccc}
\hline
\multirow{2}{*}{} & \multicolumn{2}{c}{\textbf{Our model}} & \multicolumn{2}{c}{\textbf{Base (Llama 3 8B 4-bit)}} \\ \cline{2-5} 
                  & 0-shot     & 5-shot     & 0-shot      & 5-shot     \\ \hline
MMLU              &  \textbf{0.3506}         &   0.3462 ($-$1.25\%)         & 0.6056            &   \textbf{0.6340} ($+$4.69\%)        \\  \hline
ARC-Easy              &  0.6271          & \textbf{0.7020} ($+$11.94\%)           & 0.7950            &  \textbf{0.8346} ($+$4.98\%)          \\
ARC-Challenge              &  0.3183          & \textbf{0.3797} ($+$19.29\%)           &  0.5017           & \textbf{0.5179}  ($+$3.23\%)          \\ \hline
Winogrande              & 0.5801           & \textbf{0.6275} ($+$8.17\%)          & 0.7340      &       \textbf{0.7561}  ($+$3.01\%)        \\ \hline
TruthfulQA MC1              & 0.2827           &   -         &  0.2656           &   -         \\ 
TruthfulQA MC2              &  0.4351          &    -        &  0.4305           &   -         \\  \hline
\end{tabular}
\captionof{table}{Our model v/s the base model on English Benchmarks. As expected, the domain adaptation has caused forgetting. The \% change in the scores in 5-shot runs compared to 0-shot runs are also provided. The greater improvements in the 5-shot runs show possible latent retention.}
\end{table*}

\subsubsection{Attention pooling for word tokens}
Since Nepali is not officially supported by the Llama 3 tokenizer, the token fertility of Nepali is high. This should be true for many other South Asian languages as well. The study of the attention maps is complicated by this because higher the tokens per word the more difficult it is to map attention between the tokens. Higher fertility not only complicates evaluation, but also makes inference and training costly.

To address this issue, we experimented with methods to pool the token attentions in order to construct word attentions. We applied max-pooling and mean-pooling. For max-pooling, for every Nepali word we take the element-wise max between the vectors corresponding to each constituent token to get the word attention. For mean-pooling, we take the element-wise mean.

Our experiments show max-pooling to be more suitable. We found mean-pooling normalizes attention weights to a great degree, decreasing variance. Thus, for our studies, we max-pool the token attentions to get word attention.

\subsection{Questions}
For \textbf{Q1}, we prompt the base and final models with a set of Nepali questions to generate answers. We then use GPT-4o to score these responses. Automatic evaluation of LM generations has been used with good results due to the multilinguality of frontier language models. GPT4 and GPT-4o perform well even in languages they do not officially support, Nepali included \cite{openai2024gpt4technicalreport, romanou2024includeevaluatingmultilinguallanguage, hada2024largelanguagemodelbasedevaluators}. We let GPT-4o score the answers on different qualities on scales of 0-10. We analyze the score distributions to answer Q1.

For \textbf{Q2}, we use LM Evaluation Harness to evaluate the performance of both the models on several English benchmarks and study how the scores change, or do not. This gives us insight into the forgetting in the final model. Though the base model was trained on eight languages, we only focus on its retention of English-language knowledge. We evaluate the model on MMLU \cite{hendrycks2021measuringmassivemultitasklanguage}, ARC \cite{clark2018thinksolvedquestionanswering}, Winogrande \cite{sakaguchi2019winograndeadversarialwinogradschema}, and TruthfulQA \cite{lin2022truthfulqameasuringmodelsmimic} benchmarks.

\textbf{Q3.} Dependency relations are an important feature of languages. The ability of a language model to resolve a language can be studied by analyzing the layer-head attentions of the model. We use BertViz to analyze the models at the layer- and attention head--level to accomplish this.

We curate Nepali sentences focusing on adjectives and pronouns to study how the layers in the final model encode the information about dependency relation in the sentences. We visualize self-attention in the model.

\section {Results} 
To answer \textbf{Q1}, we evaluate text generated by the base model and our model based on five attributes: correctness, grammatical correctness, usability, hallucination tendency, and overall quality.

First, we prompt both the models to answers 78 Nepali questions extracted from a traffic license exam in Nepali. Once we have the generated outputs, we let GPT-4o grade each generation on a scale of 0--10, for all five attributes.

The score distributions in the charts show the distinction between the two models. The base model (Figure 1a) shows a heavy concentration of scores at 0-1 across all metrics. This suggests that the base model's Nepali generation abilities are limited.

Our model has a more balanced score distribution (Figure 1b). While some generations still receive low scores, we observe higher scores overall compared to the base model. This is specifically evident in the scores for grammatical correctness. Our model shows strong performance here, with many generations scoring 8 or above, suggesting the model has learned how Nepali sentence are structured.

Hallucination scores demonstrate that our model has a higher median compared to the base model. This seems counterintuitive given higher hallucination is a bad quality for a language model to have. But it also suggests that our model's generations contain content that is more verifiable and can be assessed for hallucination, whereas the base model's outputs may be too limited or generic to evaluate factual accuracy.

Both box-plots (Figure 1c and 1d) confirm these observations, evidenced by broader distributions and higher medians for the final model across all metrics.

These results show that our model achieves improvements over the base model across all evaluated dimensions. The broader distribution suggests that our model is capable of generating more sophisticated and varied responses, even though this comes with some increased variability in performance.

For \textbf{Q2}, we evaluate the final model on popular English benchmarks in order to identify whether it was able to retain its knowledge in English post-pretraining. The scores of our model versus the base model in the selected benchmarks are reported in Table 1. On MMLU, our model scores 0.3506 and 0.3462 for 0-shot and 5-shot settings respectively. The base model scores 0.6056 and 0.6340 respectively, which suggests some forgetting has taken place.

On ARC-Easy, our model achieves scores of 0.6271 (0-shot) and 0.7020 (5-shot), while showing lower performance on the more challenging ARC-Challenge subset with scores of 0.3183 and 0.3797 for 0-shot and 5-shot settings respectively. The base model unsurprisingly scores higher on both benchmarks.

On the Winogrande benchmark, our model scores 0.5691 (0-shot) and 0.6022 (5-shot). For the TruthfulQA evaluation, our model achieves scores of 0.2607 and 0.4243 on MC1 and MC2 variants respectively, showing comparable performance to the baseline's 0.2656 and 0.4305.

With these numbers, it is easy to establish that forgetting has happened. However, it is noteworthy that 5-shot prompting over 0-shot generally yields higher percent increase for our model than the base model, suggesting that our model leverages few-shot examples more effectively than the final model. The highest increase in performance is for the ARC-Challenge dataset where we see a 19.29\% performance increase in the 5-shot setting compared to 0-shot. This might suggest that if properly pretrained, forgetting can be curtailed by increasing shots while prompting.

\begin{figure*}[!tb]
  \centering
  \subfloat[Base model on English adjectives.]{\includegraphics[width=0.5\textwidth]{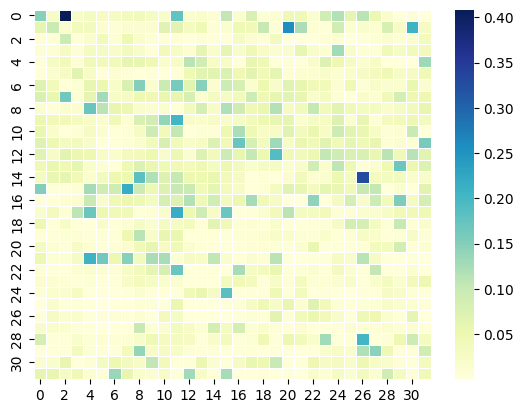}\label{fig:f5}}
    \hfill
  \subfloat[Base model on Nepali adjectives.]{\includegraphics[width=0.5\textwidth]{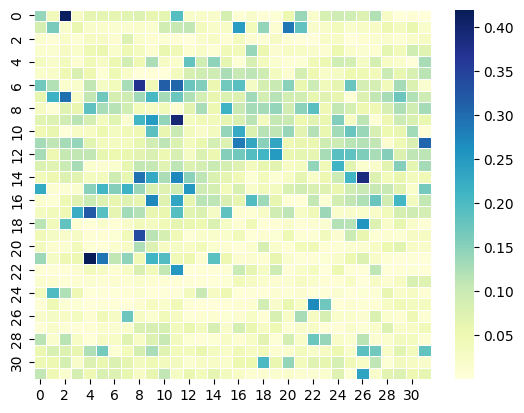}\label{fig:f6}}
    \vfill
  \subfloat[Our model on English adjectives.]{\includegraphics[width=0.5\textwidth]{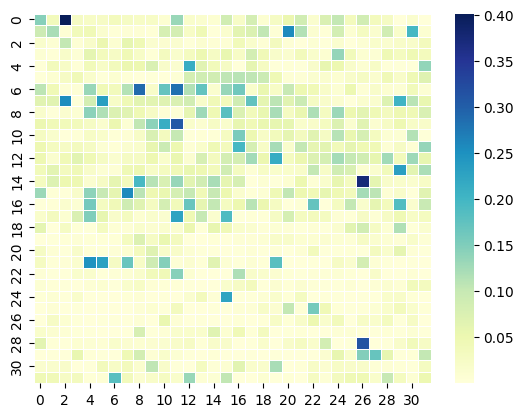}\label{fig:f7}}
    \hfill
  \subfloat[Our model on Nepali adjectives.]{\includegraphics[width=0.5\textwidth]{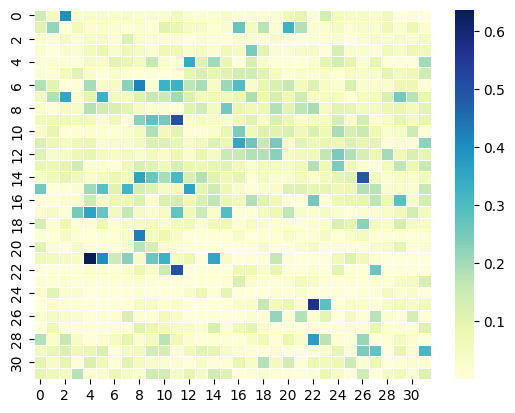}\label{fig:f8}}
  \caption{Layer-head heatmaps visualizing attention from adjectives to their respective nouns in English (a,c) and Nepali (b,d) for the base model (a,b) our model (c,d). Rows are layers and columns are attention heads. From b) and d), we can see our model has learned to attend to Nepali adjectives the way the base model attends to English ones in a).}
\end{figure*}

Finally, to answer \textbf{Q3}, we annotate a set of Nepali sentences by mapping adjectives to corresponding nouns. We explore the dependency resolution ability of the model by analyzing the attentions from the adjectives to their respective nouns across all attention heads in all layers. For each (adjective, noun) pair we extract attentions across all attention heads and find the mean of such attention heatmaps for multiple samples to get an \textit{adjective concept}. For English we average the heatmaps from 17 adjective-noun pairs and for Nepali 26 pairs. We compare the heatmaps for the base model and the final model to establish whether the final model actually captures some understanding of the language that was not present in the base model. In Figure 2 the heatmaps visualize self-attention patterns across the 32 layers (y-axis) and the 32 attention heads (x-axis) of the models. The darker blue colors indicate stronger attention weights.

Comparing our model's attention heatmaps (c,d) with the base model's heatmaps (a,b), we observe that our model has learned to process Nepali adjectives in a manner very similar to how the base model processes English adjectives. This is evidenced by the sparser and focused attention patterns in (d) as compared to the more diffuse patterns in (b). This alignment suggests improved cross-lingual transfer during pretraining. As suggested in other studies \cite{liu-etal-2019-linguistic, vig-belinkov-2019-analyzing}, we found that some of the most prominent attention heads are located in the middle layers.

The models have very different attention patterns in the lower layers (1-8), indicating that language-specific processing is perhaps performed in the earlier layers of the network. The attention patterns for English adjectives (a,c) are similar between the two models, which suggests that the DAPT only impacted the processing of Nepali in the model without disturbing its understanding of English structures.

\section{Conclusion}
We explored the utility of the continual learning paradigm in low-resource tasks, with a focus on the Nepali language. We experimented with the Llama 3 8B model to establish a simple and intuitive pretraining procedure, followed by mixed-language fine-tuning. We used automatic evaluation to grade model responses and established that the model after DAPT can generate semantically correct Nepali. We performed evaluations with several benchmarks to gauge the forgetting in the model. We finally investigated attention heatmaps to evaluate the model's grammatical knowledge in Nepali. By adapting a pretrained model to the Nepali language using only synthetic data and very limited resources and establishing generation abilities and linguistic knowledge in the new model, we make a case for domain-adaptive pretraining as a meaningful direction to explore for data- and resource-constrained languages.

\section{Limitations}
This work focuses on resource-constrained domain adaptation. Experiments are performed in a quantized 4-bit setting and the data used is synthetically generated. Pretraining sessions were run only for a single epoch and the data is mostly from online news sources, which we conjecture lead to more hallucination. Resource constraints are therefore the biggest limitation of this work. Second, we use GPT-4o for evaluation of model output. While auto-evaluation is becoming widely-adopted in multilingual research, use of human evaluators (especially domain experts for Nepali) could lead to a more definitive assessment. Similarly, there are no LM benchmarks in Nepali, which could have helped with the evaluation.

A possible extension of this work could be to study how other low-resourced languages in South Asia respond to these methods. It would also be interesting to investigate if transfer from another Indic language (opposed to English) would yield different results.

\bibliography{acl_latex}
\end{document}